%% file: main.tex
\documentclass[sigconf, 9pt]{acmart}

\input{package.tex}

\copyrightyear{2026}
\acmYear{2026}
\setcopyright{cc}
\setcctype{by}
\acmConference[ICCAD '26]{IEEE/ACM International Conference on Computer-Aided Design}{November 08--12, 2026}{San Jose, CA, USA}
\acmBooktitle{IEEE/ACM International Conference on Computer-Aided Design (ICCAD '26), November 08--12, 2026, San Jose, CA, USA}
\acmDOI{10.1145/3831252.3833989}
\acmISBN{979-8-4007-2873-0/2026/11}

\begin{document}

\input{section/0-abstract}

\title{CorrectRTL: Boost RTL Generation Quality Through \\ Automated Self-Correction with Instruction Decomposition} 

\author{Jing Wang, Shang Liu, Hangan Zhou, Wenji Fang, Yuchao Wu, Yugao Zhu, Zhiyao Xie\textsuperscript{*}}
\affiliation{%
  \institution{Hong Kong University of Science and Technology}
  \city{Clear Water Bay}
  \country{Hong Kong}
}
\affiliation{%
  \institution{\textnormal{\texttt{\{jwangjw,sliudx,hzhoubu,wfang838,ywu092,yzhuel\}@connect.ust.hk, eezhiyao@ust.hk}}}
  \country{}
}

\maketitle

\let\oldthefootnote\thefootnote
\newcounter{tempcnt}
\setcounter{tempcnt}{\value{footnote}}

\renewcommand{\thefootnote}{\fnsymbol{footnote}}
\setcounter{footnote}{1}
\footnotetext{Corresponding author.} 


\renewcommand{\thefootnote}{\oldthefootnote}
\setcounter{footnote}{\value{tempcnt}}




\input{section/1-intro}

\input{section/2-preliminaries}

\input{section/3-method}

\input{section/4-experiments}

\input{section/6-conclusion}

\bibliographystyle{ACM-Reference-Format}
\bibliography{ref}

\end{document}

%% file: package.tex
\usepackage{hhline}
\usepackage{colortbl}
\usepackage{centernot}
\usepackage{pbox}
\usepackage{amsmath}
\usepackage{amsfonts}
\usepackage{url}
\usepackage{bm}
\usepackage{comment}
\usepackage{graphicx}
\usepackage{subfigure}
\usepackage{epstopdf}
\usepackage{multirow}
\usepackage{color}
\usepackage{tabu}
\usepackage{mdframed}
\usepackage{syntax}
\usepackage{fancyvrb}

\usepackage[noend]{algpseudocode}
\usepackage{algorithmicx}
\usepackage{algpseudocode}
\usepackage{paralist}
\usepackage{stmaryrd}
\usepackage{tikz}
\usepackage[referable]{threeparttablex}
\usepackage{tabularx}
\usepackage{booktabs}
\usepackage{array}
\usepackage{fancyhdr}
\usepackage{float}
\usepackage{xspace}
\usepackage{pifont}
\usepackage{balance}
\usepackage{xcolor}

\usepackage{tablefootnote}
\usepackage[normalem]{ulem}

\usepackage{tikz}
\usepackage{xcolor}

\usepackage{graphicx}
\usepackage{adjustbox}

\usepackage{float}

\newfloat{figtab}{htb}{fgtb}
\makeatletter
  \newcommand\figcaption{\def\@captype{figure}\caption}
  \newcommand\tabcaption{\def\@captype{table}\caption}
\makeatother

\definecolor{princetonorange}{RGB}{255,143,0}

\definecolor{lightgreen}{RGB}{198, 224, 183}

\definecolor{lightred}{RGB}{240, 205, 176}

\usepackage{geometry}
 \definecolor{softblue}{rgb}{0.18,0.46,0.78}    
\definecolor{softgreen}{rgb}{0.03,0.60,0.27}

\usepackage[ruled,vlined,linesnumbered]{algorithm2e}

\usepackage{dblfloatfix}

\definecolor{newblue}{RGB}{66, 173, 245}
\definecolor{cyan}{RGB}{213,229,255}
\definecolor{yellow}{RGB}{253, 243, 208}
\definecolor{orangeShallow}{RGB}{255,190,0}
\definecolor{ShallowYellow}{RGB}{249,241,204}
\definecolor{ShallowGreen}{RGB}{222,237,214}
\definecolor{ShallowOrange}{RGB}{250,229,212}
\definecolor{NormalGreen}{RGB}{51, 105, 30}
\definecolor{ShallowPurple}{RGB}{232, 223, 243}

\definecolor{ablationorange}{RGB}{255, 245, 220}

\definecolor{ablationpurple}{RGB}{250, 240, 245}

\definecolor{ablationblue}{RGB}{235, 245, 255}

\definecolor{ablationgreen}{RGB}{225, 250, 235}

\definecolor{ablationgray}{RGB}{231, 231, 231}

%% file: section/0-abstract.tex
\begin{abstract}

Large language models (LLMs) have shown strong promise in specification-to-RTL generation. Among the explorations, the \emph{self-correction paradigm} has significantly advanced LLM-based RTL generation. However, two challenges limit further progress. \textbf{Challenge on benchmarks.} Existing benchmarks are saturated by advanced LLMs, leaving little room to evaluate the agentic self-correction framework. \textbf{Challenge on methodology.}~The \emph{monolithic self-correction} paradigm rewrites the entire design at once, where LLMs are susceptible to error confusion and corruption of originally correct code.
To address the benchmark challenge, we contribute \textbf{ScaleBench}, a more challenging benchmark of 200 complex RTL designs.
To address the challenge on methodology, we propose \textbf{CorrectRTL}, an agentic framework that realizes a new \textbf{decompose-then-correct} paradigm, in which the LLM agent explicitly decouples scaffold construction from detailed implementation. \emph{CorrectRTL} contributes three core techniques: (1)~\emph{Structure-oriented decomposition}, (2)~\emph{Snippet-based RAG}, and (3)~\emph{SVA-based self-correction}.
On ScaleBench, CorrectRTL improves Pass@1 from 20.5\% to 49.0\% (+28.5\%). On VerilogEval, CorrectRTL achieves up to 96.1\% Pass@1, demonstrating consistent improvements across both simple and complex designs.

\end{abstract}

%% file: section/1-intro.tex

\section{Introduction}\label{sec:intro}

Large language models (LLMs) are rapidly transforming hardware design automation, from RTL generation~\cite{pearce2020dave, thakur2024verigen, chang2023chipgpt, liu2024rtlcoder, liu2025deeprtl} to verification and debugging~\cite{kang2024fveval, yao2024hdldebugger, tsai2024rtlfixer, fang2024assertllm}. Among the various stages of the design flow, specification-to-RTL generation has attracted intensive attention~\cite{liu2023verilogeval, lu2024rtllm, pinckney2024revisiting, zhao2024mage, ho2024verilogcoder}. LLMs can understand hardware specifications and produce synthesizable designs. Such LLM-assisted design flows are emerging as a compelling paradigm that promises to fundamentally reshape hardware development. 

Recently, a new paradigm~\cite{zhao2024mage, ho2024verilogcoder} achieves promising results on VerilogEval~\cite{liu2023verilogeval}, one of the most commonly used RTL generation benchmarks. We refer to this paradigm as the \emph{self-correction} paradigm, in which the model first generates RTL code and then corrects it based on verification feedback. Specifically, MAGE~\cite{zhao2024mage} and VerilogCoder~\cite{ho2024verilogcoder} achieve pass rates exceeding 94\%, approaching near-perfect functional correctness. These two SOTA works show a more promising direction for automated RTL generation than other specification-based prompt engineering strategies. One important reason is that simple prompt engineering on specifications for LLMs is inherently hard to address detailed signal-level bugs.
However, the existing \emph{self-correction} works~\cite{zhao2024mage, ho2024verilogcoder} rely on manually written golden testbenches~\cite{ho2024verilogcoder, zhao2024mage}. In real applications, manually crafting golden testbenches for every module is impractical. 
To further advance the self-correction paradigm toward greater automation, some works begin to explore auto-generated testbenches~\cite{tan2025autoverifix, huang2024towards, sami2024aivril, tsai2024rtlfixer} or Python reference models~\cite{tan2025autoverifix}. However, two important challenges significantly limit the effectiveness and further advancement of the research direction.

\begin{figure}[!t]
    \centering
    \includegraphics[width=0.95\linewidth]{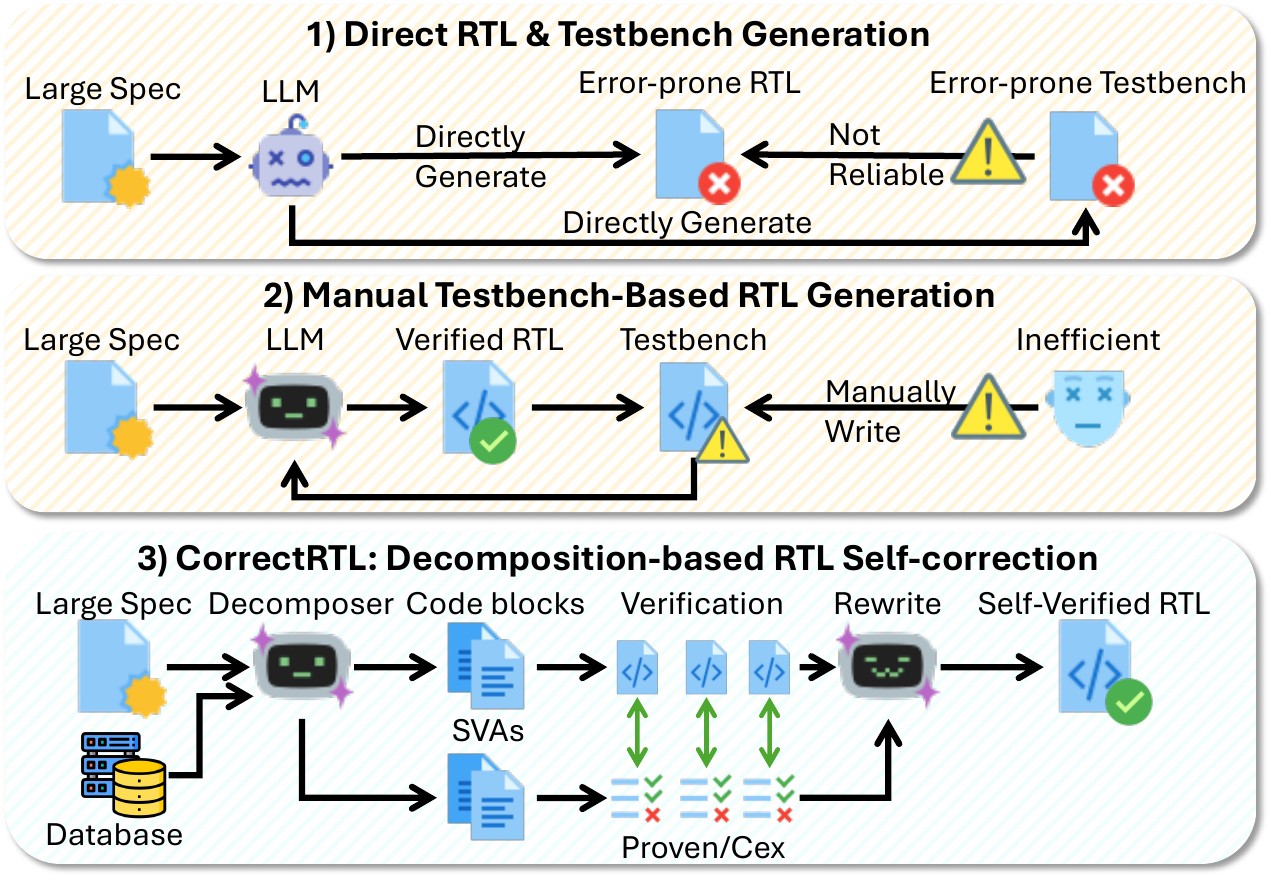}
        \vspace{-.15in}
    \caption{Overview of CorrectRTL. (1) Existing approaches rely on testbenches or reference code for self-correction, which are unreliable or impractical. (2) CorrectRTL combines structure-oriented decomposition, snippet-based RAG, and SVA-based self-correction to improve both generation quality and error identification.}
    \label{fig:teaser-correctrtl}
    \vspace{-.3in}
\end{figure}

\textbf{Challenge on benchmarks.}
Existing benchmarks such as VerilogEval~\cite{liu2023verilogeval} and RTLLM~\cite{lu2024rtllm} consist predominantly of simple designs. Direct LLM generation already achieves over 80\% pass rate on these benchmarks, and the remaining failures are dominated by a small set of recurring error patterns. This leaves little room to differentiate methods or to evaluate whether \emph{self-correction} techniques genuinely generalize, rather than merely fixing a narrow class of easy errors. A more challenging benchmark with more complex designs is needed to meaningfully assess the effectiveness of self-correction approaches.

\textbf{Challenge on methodology.} We refer to the current approach as the \emph{monolithic self-correction} paradigm, in which LLMs generate a full verification reference (e.g., a testbench or a Python reference model) and apply corrections to the entire design as a whole. This paradigm faces two fundamental difficulties. \textbf{Difficulty 1: reference difficulty}. The correctness of the verification reference itself is inherently hard to guarantee. Auto-generated testbenches or reference code are not inherently more reliable than the RTL they are meant to verify. Generating a complete testbench or reference model becomes increasingly challenging as design scale increases.
\textbf{Difficulty 2: correction effectiveness}. Monolithic correction forces the LLM to simultaneously reason about the global structure and local behavioral logic. This entanglement of the whole design leads to two common failure reasons: (1)~The model becomes \emph{confused} by the redundant context and fails to identify the actual errors. (2)Even if the error is correctly localized, the model \emph{corrupts originally correct code} while attempting to fix the error, because it inadvertently modifies the surrounding code. As design complexity grows, these failure modes compound, making monolithic self-correction increasingly unreliable.

To address the above challenges, we make two contributions. \textbf{(1) New benchmark.} We propose \textbf{ScaleBench}, a benchmark containing larger and more complex designs that are substantially more challenging than those in existing benchmarks~\cite{lu2024rtllm, liu2023verilogeval, pinckney2024revisiting}. 
\textbf{(2) New paradigm.} We propose a new self-correction paradigm. For \textbf{reference}, we propose using SVA-based verification because SVA is lightweight, thus inherently shorter and easier to generate. Generating SVAs will not become more challenging with the design scaling up. For \textbf{correction effectiveness}, we propose to \emph{decompose and correct}. This paradigm decouples structural scaffold construction from behavioral detail implementation. We implement an agentic framework, \textbf{CorrectRTL}, to instantiate and validate this paradigm, as depicted in Figure~\ref{fig:teaser-correctrtl}.
\emph{CorrectRTL} involves three key techniques to realize our paradigm.\footnote{The benchmark and dataset will be open-sourced at \url{https://github.com/hkust-zhiyao/CorrectRTL}.}

(1)~\textbf{Structure-oriented decomposition.} We observe that LLMs are remarkably capable of constructing the \emph{structural scaffold} of a design, such as declaring modules, defining port interfaces, and organizing sequential and combinational blocks. But LLMs frequently fail at \emph{behavioral details} within individual blocks. This asymmetry motivates us to explicitly decouple scaffold construction from detail implementation. So that LLMs first produce a well-organized scaffold via specification decomposition into block-level instructions, and self-correction is then focused exclusively on detail errors within individual blocks without disrupting the overall architecture.

(2)~\textbf{Snippet-based RAG.} We further observe that LLMs are often \emph{stubborn} on detail-level errors. General skill injection through prompt engineering or high-level design guidelines fails to correct many recurring mistakes. Because the model lacks the precise, context-specific knowledge needed for each logic block. This motivates us to design a RAG system that provides exact snippet-level examples of correct implementations for the specific behavioral patterns the LLM struggles with. Our decomposition naturally serves this purpose. Specifically, the decomposed instruction can serve as a query to retrieve relevant examples.

(3)~\textbf{SVA-based self-correction.} We propose using lightweight SystemVerilog Assertions (SVAs) that target a single logic property of an individual block. Each SVA checks one focused property, so it is easier to generate correctly than a full golden testbench. An EDA tool checks the SVA against the block, and any counterexample pinpoints the faulty block and drives a targeted rewrite. This enables block-level error identification and correction without a manually written golden testbench.

Our contributions are summarized as follows:
\begin{itemize}
    \item \textbf{ScaleBench benchmark.} We propose ScaleBench, a challenging benchmark of 200 more complex RTL designs. 
    \item \textbf{Decompose-and-correct paradigm.} We propose a new self-correction paradigm that explicitly decouples structural scaffold construction from behavioral detail implementation. 
    \item \textbf{CorrectRTL framework.} We implement an agentic framework that instantiates the decompose-and-correct paradigm. 
    \item \textbf{Snippet-level RAG dataset.} We construct a large-scale RTL-specific RAG dataset, enabling fine-grained knowledge injection at the block level.
\end{itemize}

%% file: section/2-preliminaries.tex
\begin{table}[t!]
    \centering
    \resizebox{\linewidth}{!}{%
    \begin{tabular}{l||c|c|c|c|c}
    \hline
    \multirow{2}{*}{\textbf{Method}} & \textbf{Instr.} & \textbf{SVA-based} & \textbf{Self-corr.} & \textbf{Snippet} & \textbf{Correct} \\
    & \textbf{decomp.} & \textbf{ident.} & \textbf{w/o golden TB} & \textbf{RAG} & \textbf{func.} \\
    \hline
    \hline
    \multicolumn{6}{l}{\textit{Syntax error correction only}} \\
    \hline
    HDLDebugger~\cite{yao2024hdldebugger} &   & &   & & \\
    RTLFixer~\cite{tsai2024rtlfixer} &   & &   & & \\
    \hline
    \multicolumn{6}{l}{\textit{Self-corr. relying on manually written golden testbench}} \\
    \hline
    MAGE~\cite{zhao2024mage}  & & & & & \checkmark  \\
    VerilogCoder~\cite{ho2024verilogcoder} & \checkmark &    &  & & \checkmark  \\
    \hline
    \multicolumn{6}{l}{\textit{Self-corr. without manually written golden testbench}} \\
    \midrule
    \rowcolor{softblue!12} VeriAssist~\cite{huang2024towards} & & & \checkmark & & \checkmark   \\
    \midrule
    \rowcolor{softblue!12} AutoVeriFix~\cite{tan2025autoverifix} &  &  & \checkmark & & \checkmark  \\
    \midrule
    \rowcolor{softgreen!14}\textbf{CorrectRTL} (ours) &
      \checkmark & \checkmark &
      \checkmark & \checkmark & \checkmark \\
    \bottomrule
    \end{tabular}}
    \caption{Comparison of related works. TB = testbench. CorrectRTL is the first to combine decomposition, SVA-based identification, and snippet-level RAG for RTL generation.}
    \label{tab:related-works}
\end{table}

\section{Preliminary}

\subsection{Related Work}\label{sec:pre}

In this section, we mainly focus on automated self-correction on RTL generation~\cite{huang2024towards, tan2025autoverifix, sami2024eda, huang2024towards, yao2024hdldebugger}. 
Table~\ref{tab:related-works} summarizes the related works in the scope.
Among these works, some very early works explored syntax identification~\cite{yao2024hdldebugger, tsai2024rtlfixer}, which is not our focus. 
We separate the self-correction works on RTL into two main types: (1) \emph{self-correction} relying on a manually written golden testbench, and (2) \emph{self-correction} without a manually written golden testbench.

\textbf{Self-correction without golden testbench} is the most challenging yet meaningful direction. Existing works explore automated testbench generation~\cite{huang2024towards} and Python reference code~\cite{tan2025autoverifix}. However, both face fundamental reliability issues: auto-generated testbenches are harder to guarantee correct than RTL itself, and Python reference models remain challenging for precise hardware description. Critically, both approaches focus on the \emph{verification} side---they cannot inject domain knowledge that the LLM lacks.

\textbf{RAG for code generation.} Retrieval-augmented generation has been widely applied in software code generation~\cite{ragsurvey2024}, but remains unexplored for RTL. Standard code RAG retrieves whole files or functions, which is too coarse-grained for hardware designs where a single incorrect bit width or timing polarity causes functional failure. CorrectRTL introduces \emph{snippet-level RAG} specifically designed for RTL: decomposed code blocks paired with specification sections, retrieved via a hardware-aware hybrid retriever.

CorrectRTL takes a fundamentally different approach: rather than improving verification feedback alone, we \emph{inject design-family knowledge} into the generation process via snippet-based RAG, while using SVA-based self-correction as a complementary error identification mechanism. The structure-oriented decomposition serves as the architectural backbone, enabling both fine-grained generation and RAG database construction.

%% file: section/3-method.tex
\begin{table}[t]
    \centering
    \small
    \begin{tabularx}{\linewidth}{@{}c>{\raggedright\arraybackslash}X@{}}
    \hline
    \textbf{Symbol} & \textbf{Meaning} \\
    \hline
    $\mathcal{S}$ & Design specification (natural language) \\
    $f_\theta$ & LLM with parameters $\theta$ \\
    $\mathcal{C}$,\; $\mathcal{C}'$ & Generated / corrected RTL code \\
    $\mathcal{C}_i$,\; $\mathcal{C}_i'$ & RTL code block $i$ (generated / corrected) \\
    $\mathcal{I}_i^\text{R}$,\; $\mathcal{I}_i^\text{A}$ & RTL / SVA instruction for block $i$ \\
    $\mathcal{A}_i$ & SVA generated for block $i$ \\
    $\mathcal{D}$ & RAG snippet database \\
    $\textsc{Retrieve}(q, \mathcal{D})$ & Dense semantic retriever; returns top-$K$ pairs \\
    $\mathcal{T} = (\mathcal{V}, \mathcal{E})$ & Hierarchical design structure tree \\
    $\mathcal{V}_\text{mod}$,\; $\mathcal{V}_\text{blk}$ & Module / block nodes in $\mathcal{T}$ \\
    $\textsc{Parse}(\mathcal{C})$ & Structural parser; extracts blocks from RTL \\
    \hline
    \end{tabularx}
    \caption{Notation used in this paper.}
    \label{tab:notation}
    \vspace{-.2in}
\end{table}

\subsection{Problem Formulation}

Table~\ref{tab:notation} summarizes the notation. LLM-based RTL generation with self-correction follows a \emph{generate-then-correct} paradigm. Given a design specification $\mathcal{S}$, an LLM $f_\theta$ produces RTL code $\mathcal{C}$ and a verification artifact $\mathcal{A}$:
\begin{align}
    \mathcal{C} = f_\theta(\mathcal{S}), \qquad \mathcal{A} = f_\theta(\mathcal{S})
\end{align}
An EDA tool checks $\mathcal{C}$ against $\mathcal{A}$, yielding error feedback $\mathcal{F} = \textsc{Verify}(\mathcal{C}, \mathcal{A})$, which guides correction: $\mathcal{C}' = f_\theta(\mathcal{C}, \mathcal{F}, \mathcal{A})$.

The verification artifact $\mathcal{A}$ differentiates existing self-correction paradigms: golden testbenches~\cite{ho2024verilogcoder, zhao2024mage} (effective but impractical at scale), auto-generated testbenches~\cite{huang2024towards} (harder to guarantee correct than RTL itself), Python reference models~\cite{tan2025autoverifix} (challenging for precise hardware semantics), and \emph{verification SVAs} (ours). Each SVA targets a single logic property, making it easier to generate correctly than a full testbench.

Regardless of the choice of $\mathcal{A}$, generation quality is bounded by $f_\theta$'s domain knowledge. CorrectRTL addresses this by injecting verified design-family knowledge from a curated RAG database $\mathcal{D}$ into both RTL and SVA generation:
\begin{align}
    \mathcal{C} = f_\theta\!\big(\mathcal{S},\; \textsc{Retrieve}(\mathcal{S}, \mathcal{D})\big)
\end{align}
This improves both phases at once. It produces better RTL in Phase~I and better SVAs in Phase~II, rather than relying on verification feedback alone. \textcolor{black}{Our paradigm can boost the performance on both generation and correction steps.}

\section{Benchmark and Dataset}
In this section, we introduce the new challenging specification-to-RTL benchmark in ~\ref{sec:scalebench} and the RAG dataset in Section~\ref{sec:rag-database}.
\subsection{ScaleBench: A Challenging RTL Benchmark}\label{sec:scalebench}

Existing benchmarks are insufficient for evaluating self-correction methods on complex designs. To address the limitation and evaluate LLMs' ability more thoroughly, we propose \textbf{ScaleBench} with 200 larger RTL designs. Table~\ref{tab:scalebench-stats} compares ScaleBench with existing benchmarks in terms of scale and difficulty. Table~\ref{tab:scalebench-stats} demonstrates the comparison of ScaleBench with existing benchmarks. Our ScaleBench benchmark contains significantly larger (Avg. 120 LoC compared to RTLLM's 48 and VerilogEval's 16 LoC). The lower performance of the baseline (Claude-4.5 Sonnet) also underscores the benchmark's challenging nature.
To curate the benchmark, we select 200 designs from the most recent work RTLPP~\cite{rtlpp2025} that proposes a high-quality dataset with aligned specifications and RTL pairs. We manually verify the quality of the specifications and make improvements.

\begin{table}[t]
\centering
\small
\begin{tabular}{l|c|c|c|c}
\hline
\textbf{Benchmark} & \textbf{\#Designs} & \textbf{Avg LoC} & \textbf{Max LoC} & \textbf{Baseline} \\
\hline
VerilogEval~\cite{liu2023verilogeval} & 156 & 16 & 64 & 80.1\% \\
RTLLM~\cite{lu2024rtllm} & 50 & 48 & 175 & 68.0\% \\
\textbf{ScaleBench} & \textbf{200} & \textbf{120} & \textbf{348} & \textbf{20.5\%} \\
\hline
\end{tabular}
\scriptsize Baseline pass rates are from Claude Sonnet 4.5 direct generation.
\caption{Benchmark comparison.  ScaleBench provides larger designs that are more challenging for LLMs.}
\vspace{-.2in}
\label{tab:scalebench-stats}
\end{table}

\begin{table}[b]
\centering\small
\setlength{\tabcolsep}{4pt}
\begin{tabular}{l|c|l}
\toprule
\multicolumn{3}{c}{\textbf{Code snippets (by structural block type)}} \\
\midrule
Block type & Snippets & Target logic \\
\midrule
Sequential block  & $6{,}092$ & registers, counters, FSM state \\
Assign group      & $2{,}218$ & control fan-out, datapath \\
Combinational block & $936$   & muxes, decoders, arithmetic \\
\textbf{Total}    & $\mathbf{9{,}246}$ & \\
\midrule
\multicolumn{3}{c}{\textbf{SVA templates (by invariant family)}} \\
\midrule
Invariant family & Templates & Target pattern \\
\midrule
Output equation      & $1{,}337$ & combinational output relations \\
Past-value relation  & $1{,}001$ & $X \!=\! \$past(Y)$ style timing \\
Reset initialisation & $1{,}256$ & async/sync register reset \\
Handshake stability  & $430$     & valid/ready protocols, AXI-like \\
Selector exclusion   & $315$     & mux selector $\rightarrow$ output equality \\
One-hot encoding     & $261$     & FSM state-bit invariants \\
Range / bit-width    & $260$     & bounded-signal assertions \\
Counter bounding     & $133$     & counter wrap / maximum \\
\bottomrule
\end{tabular}
\caption{Composition of the curated RAG database. Both sub-datasets share the same spec-snippet embedding key.}
\label{tab:rag-inventory}
\end{table}

\begin{figure*}[t]
    \centering
    \includegraphics[width=1\linewidth]{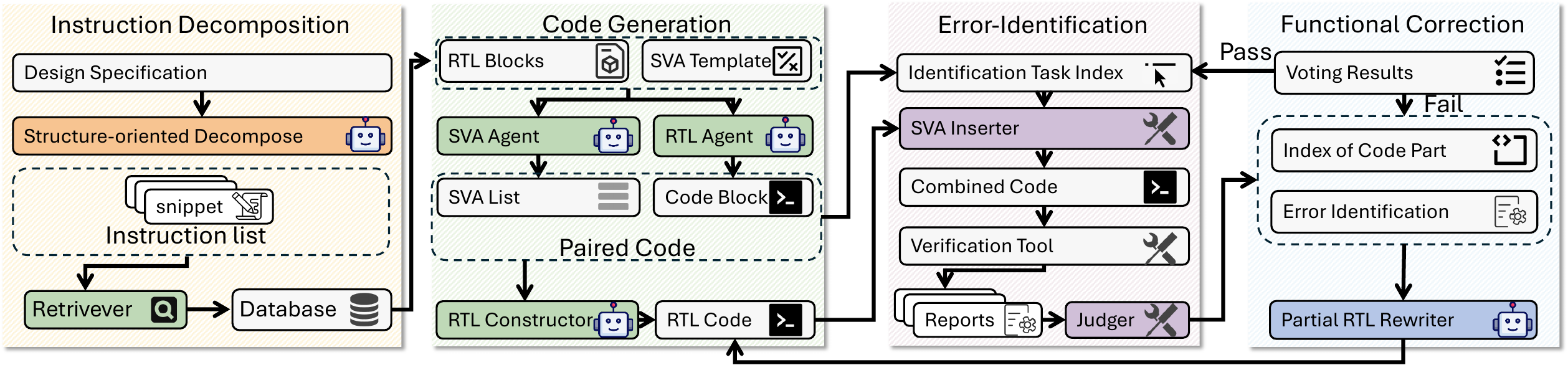}
    \caption{Overview of CorrectRTL framework, orchestrating a multi-agent system with four phases. Phase~2 is augmented with snippet-based RAG from a database constructed via the same decomposition as Phase~1.}
    \label{fig:method-workflow}
\end{figure*}

\subsection{RAG Database}\label{sec:rag-database}

General rule-based skill injection, or prompt engineering, is insufficient to augment LLM-based RTL generation for the detailed functional behaviors that LLMs often lack. To address this limitation, we curate a RAG database to provide LLM-based agents with detailed functional knowledge. We deliberately organize this database at the \emph{snippet (block) level} rather than the whole-design level, for two reasons. \emph{Reason 1.} Individual blocks---particularly their RTL bodies---are directly reusable as code exemplars, whereas whole designs mix unrelated logic and cannot be inserted into a partially-generated module. \emph{Reason 2.} A short, behavior-focused spec snippet yields a much tighter semantic signal than a full specification, so top-$k$ retrieval returns exemplars that match the queried sub-behavior rather than a loosely-related design.

\textbf{Overview.}
The RAG database has two sub-datasets that share the same retrieval key. The first is the \emph{block-level dataset}, which consists of (spec-snippet, RTL-block) pairs: each pair binds a short natural-language description to a single structural block (e.g., \texttt{always} block, \texttt{assign} group, or memory declaration) extracted from a verified golden design. The second is the \emph{SVA-template-level dataset}, which consists of formally-proven SystemVerilog assertions; each SVA template is a specific assertion that has been \emph{verified} on the golden RTL and is attached to one RTL block from the block-level dataset, so every SVA template inherits a corresponding spec-snippet and code block.
Both sub-datasets are curated on top of existing large-scale RTL training corpora (MG-Verilog~\cite{mgverilog2024}, OriGen~\cite{cui2024origen}, AutoVCoder~\cite{gao2024autovcoder}, RTLCoder~\cite{liu2024rtlcoder}, \emph{etc.}). After whitespace-normalized de-duplication, the block-level dataset contains $9{,}246$ unique (spec-snippet, RTL-block) pairs. The SVA-template-level dataset contains roughly $5$k assertions that JasperGold proves on the parent golden, spanning eight invariant families (Table~\ref{tab:rag-inventory}).

\begin{figure}[b]
    \centering
    \includegraphics[width=1.0\linewidth]{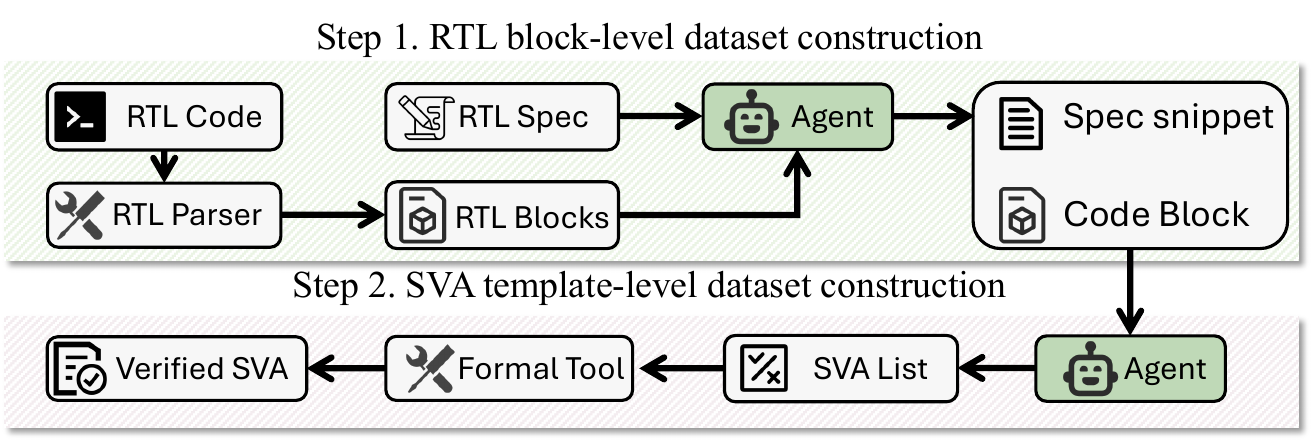}
    \caption{Two-step workflow of RAG dataset construction.}
    \label{fig:placeholder}
\end{figure}

\begin{figure}[b]
    \centering
    \includegraphics[width=1.0\linewidth]{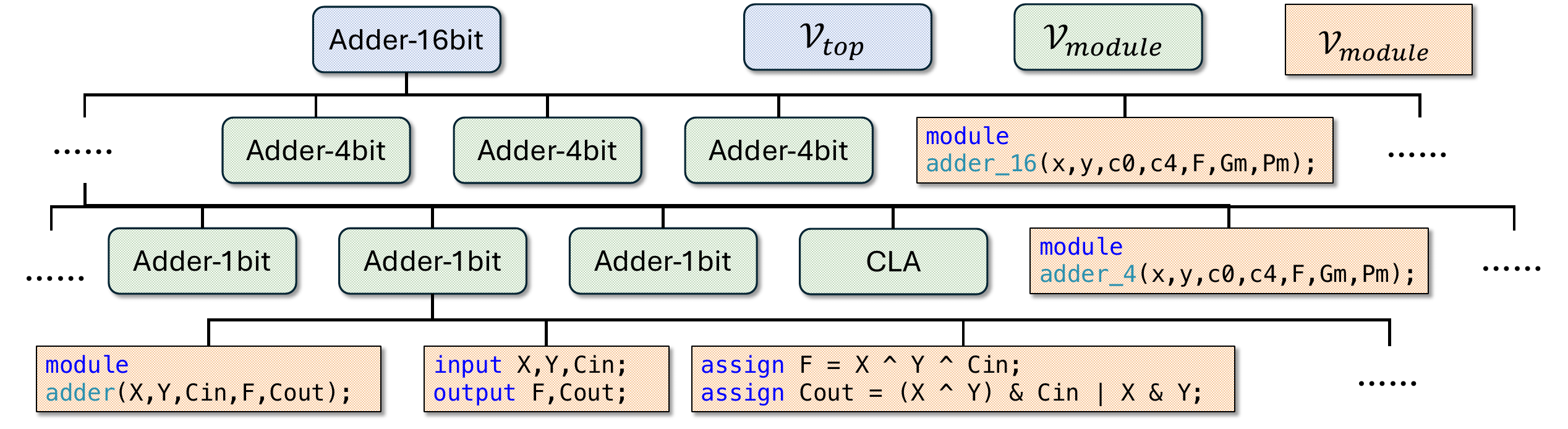}
    \caption{Hierarchical tree $\mathcal{T}$ for a 16-bit adder.}
    \label{fig:adder16-tree-structure}
\end{figure}

\textbf{Step 1: Block-level dataset curation.}
A Verilog parser decomposes each RTL module in the source corpus into its constituent structural blocks. We treat four block types: module-header / port declarations, sequential \texttt{always} blocks, combinational \texttt{always} blocks, and \texttt{assign} groups. Figure~\ref{fig:adder16-tree-structure} shows an example for a 16-bit adder, where the parser extracts the module header at the top and the per-bit and per-half adders as sequential or combinational leaf blocks below. For each block, we pass both the block code and the module's full design specification to an agent whose task is to produce a 1--3-sentence spec snippet capturing the block's behavioral semantics. Because specifications in public training corpora are often incomplete or loosely worded, the agent is not restricted to pure extraction: when the block's code exposes logic that the surrounding specification underspecifies or omits, the agent may augment the snippet with behavior directly inferred from the RTL. The resulting (spec-snippet, RTL-block) pair forms one entry of the block-level dataset.

\textbf{Step 2: SVA-template-level dataset curation.}
Building on the Step~1 output, the agent receives each (spec-snippet, RTL-block) pair and generates a set of candidate SystemVerilog assertions labeled with an invariant family (e.g., reset initialization, past-value relation, handshake stability). Each candidate is then inserted into the original golden RTL and discharged by JasperGold. Only assertions that the tool \emph{proves} on the golden are retained and become SVA templates in the database. A $1{,}000$-SVA calibration across the eight invariant families (125 per category, 855 unique designs) yields a $44\%$ overall proof rate. The per-family rate ranges from $58\%$ on the easiest family (reset initialization) to $32\%$ on the hardest (counter bounding). At the full corpus size this projects to roughly $5$k formally-verified templates.

The two sub-datasets are jointly indexed by a single dense embedding over the spec-snippet field. A single retrieval therefore returns both a reference code pattern and its verified assertion oracle for the same block family, eliminating the cold-start problem of prior RAG approaches that index arbitrary unverified code.

\section{Methodology}\label{sec:method1}

CorrectRTL assigns agents two roles. \textbf{Structural agents} decompose $\mathcal{S}$ into block-level tasks using curated rules such as module boundaries, reset conventions, and port naming. \textbf{Detail agents} generate block-level RTL and SVA using snippet-based RAG, because much domain-specific logic cannot be captured by rules alone.
Sec~\ref{sec:overview-correctrtl} gives the four-phase overview. Sec~\ref{sec:structure-decomposition} describes the decomposition strategy. Sec~\ref{sec:snippet-rag} presents RAG-augmented RTL and SVA generation. Sec~\ref{sec:functional-correction} details the correction process.

\subsection{Framework Overview}
\label{sec:overview-correctrtl}

Figure~\ref{fig:method-workflow} depicts the workflow of the CorrectRTL framework, which operates in four phases.

\textbf{Phase 1: Decomposition} (\emph{structural agents}).
Depending on design complexity, CorrectRTL applies one of two decomposition strategies. \emph{(1) Code-level decomposition}: the LLM generates the full design from $\mathcal{S}$, and structural parsing extracts blocks from the generated RTL for targeted verification. The following self-correction process is based on the block level. \emph{(2) Spec-level decomposition}: a decomposer agent extracts a hierarchical structure tree $\mathcal{T}$ from $\mathcal{S}$, and paired instructors generate RTL and SVA instruction pairs for each block. Both strategies define the block boundaries shared across generation, retrieval, and verification.

\textbf{Phase 2: RAG-Augmented RTL and SVA Generation} (\emph{detail agents}).
For each block, $\textsc{Retrieve}(q_i, \mathcal{D})$ fetches the top-$K$ (spec-snippet, code-block, proven-SVA) entries from the RAG database. The query $q_i$ is the block's spec instruction under spec-level decomposition, or an LLM-authored one-sentence description of the parsed block under code-level decomposition. One agent generates the RTL block $\mathcal{C}_i$ using the instruction and the retrieved code blocks. A second agent generates one SVA $\mathcal{A}_i$ using the same retrieval. The two agents share the retrieval so the RTL and the SVA are grounded in the same verified references. Under spec-level decomposition, a constructor assembles the blocks into the full module $\mathcal{C}$.

\textbf{Phase 3: SVA-Based Verification.}
For each block $\mathcal{C}_i$, an EDA tool checks $\mathcal{A}_i$ against $\mathcal{C}$. If the assertion produces a counterexample, the block is flagged for rewriting. SVAs target single logic properties derived from the specification, so they serve as the oracle for block-level correctness.

\textbf{Phase 4: Targeted Rewriting.}
For each flagged block, a rewriter agent reads the counterexample trace, identifies the root cause, and produces a corrected block $\mathcal{C}_i'$. Only the failing block is replaced: $\mathcal{C}' = \textsc{Replace}(\mathcal{C}, \mathcal{C}_i, \mathcal{C}_i')$. Phases~3 and 4 iterate until every block passes its SVA, with a bounded round cap to avoid oscillation.

\subsection{Structure-Oriented Decomposition}\label{sec:structure-decomposition}

Phase~1 fixes the block boundaries that all later phases use. We introduce two baselines and then propose our decomposition strategy. The first baseline is \emph{code-level decomposition}, which lets the LLM generate the whole RTL in one pass and parses blocks out of the result. The second baseline is \emph{vanilla spec decomposition}, which splits the specification into block-level instructions first and generates each block independently, with no per-type guidance. On top of vanilla spec decomposition, we propose \emph{structure-oriented decomposition}, which injects a short per-type strategy into the decomposer for five design families. CorrectRTL uses structure-oriented decomposition for its workflow. We evaluate all three as configurations in the ablation study (Sec.~\ref{sec:ablation}).

\textbf{Code-Level Decomposition.}
The LLM generates the complete RTL from the specification in one pass, and a structural parser segments the result into blocks:
\begin{align}
    \mathcal{C} = f_\theta(\mathcal{S}), \qquad \{\mathcal{C}_1, \ldots, \mathcal{C}_n\} = \textsc{Parse}(\mathcal{C}) \label{eq:code-decomp}
\end{align}
The parser recognizes structural boundaries such as \texttt{always} blocks, assign groups, and module headers. The SVA instruction $\mathcal{I}_i^\text{A}$ for each block $\mathcal{C}_i$ is produced by a single LLM call that reads $\mathcal{C}_i$ together with the portion of $\mathcal{S}$ that describes it, and writes down the key behavioral property the block must satisfy. Phase~2 then generates $\mathcal{A}_i$ from $\mathcal{I}_i^\text{A}$ and the retrieved SVA templates.

\textbf{Spec-Level Decomposition.}
The specification is first parsed into a hierarchical structure tree $\mathcal{T} = (\mathcal{V}, \mathcal{E})$ whose leaves are individual logic blocks. The block boundaries follow the same definition used in the RAG-database curation (Sec.~\ref{sec:rag-database}): the four block types are module-header / port declarations, sequential \texttt{always} blocks, combinational \texttt{always} blocks, and \texttt{assign} groups. Figure~\ref{fig:adder16-tree-structure} shows an example tree for a 16-bit adder. For each leaf block $i$, one LLM call produces two instructions in parallel. The RTL instruction $\mathcal{I}_i^\text{R}$ describes what the block should compute. The SVA instruction $\mathcal{I}_i^\text{A}$ states the key behavioral property the block must satisfy. Both instructions are grounded in $\mathcal{S}$, so the SVA instruction is independent of any later generated RTL. Phase~2 generates $\mathcal{C}_i$ and $\mathcal{A}_i$ from these two instructions, and a constructor assembles the blocks into the full module.

\textbf{Type-oriented decomposition strategies.}
Asking an LLM to decompose a specification with no extra guidance is brittle. Different design families have distinct coding conventions, and the LLM tends to produce plans that overlap across blocks, request unnecessary testbench scaffolding, or fall back to generic templates that break during assembly. We therefore condition the decomposer on a short per-type strategy. A lightweight classifier assigns each module one of five families, and the decomposer applies the strategy attached to that family before writing the block plan. For \textit{FSM} modules, the strategy splits the design into a clocked state register and a combinational next-state function, and keeps an explicit reset state so the LLM does not fall back to inferred latches. For \textit{sequential datapath} modules, it asks for one non-blocking \texttt{always} block per register group and checks that pipeline depth, counter widths, and reset values follow the spec. For \textit{combinational} modules, it restricts the plan to \texttt{assign} and \texttt{always @(*)} blocks and asks that every output be covered in every branch. For \textit{memory} modules, it pins down array dimensions, per-port read/write timing, and verbatim initialization values. For \textit{structural} modules, it asks the decomposer to list every sub-module instance, use named port connections, and declare each inter-block wire with its exact width. Each strategy targets one failure mode we observed in unconditioned decomposition, and it is relaxed when the spec itself contradicts it.

\textbf{Vanilla decomposition (baseline).}
We call direct spec-level decomposition without these skills \emph{vanilla decomposition}. An LLM is asked to break the specification into blocks and write one instruction per block, with no type-specific guidance. In practice the resulting plans are noisy. Block boundaries overlap, so two blocks end up generating the same signal. Some plans include testbench blocks or wrapper modules that the original specification does not require. FSM designs silently collapse into a single always block with inferred latches. Downstream assembly then crashes on signal redefinition or port-width mismatch. We report vanilla decomposition as a baseline in Sec.~\ref{sec:ablation} to isolate the contribution of the type-oriented skills above.

\subsection{RAG-Augmented RTL and SVA Generation}\label{sec:snippet-rag}

Generating an RTL block or an SVA from scratch is unreliable. SVA generation is especially error-prone in both syntax (wrong operators, missing sampling edges) and semantics (the property does not match the intended behavior). We therefore retrieve semantically similar examples from the RAG database and feed them to the RTL and SVA agents as references, so the agents adapt existing patterns rather than invent them.

\textbf{Transformer-based retriever.}
We use a pretrained transformer-based sentence encoder to embed every spec-snippet in $\mathcal{D}$. At inference we embed the query and return the top-$K$ nearest neighbors by cosine similarity. Each neighbor brings two things the agents can use: a reference RTL block, and a set of formally-proven SVA templates attached to that block.

\begin{table*}[!t]
\centering
    \resizebox{0.99\textwidth}{!}{
    \begin{tabular}{c|l||c|c|c|c|c|c}
    \toprule
        \textbf{Base LLM} & \textbf{Method} & \textbf{VerilogEval Human v1} & \textbf{VerilogEval Machine v1} & \textbf{VerilogEval Human v2} & \textbf{RTLLM v1.1} & \textbf{RTLLM v2} & \textbf{ScaleBench} \\
        & & (156 designs) & (143 designs) & (156 designs) & (50 designs) & (50 designs) & (200 designs) \\
    \midrule
          \multirow{5}{*}{GPT-4o~\cite{achiam2023gpt}} & Direct generation &  59.0\% &  69.2\% & 68.6\% & 69.1\% & 67.2\% & \phantom{0}9.5\% \\
          & RTLFixer~\cite{tsai2024rtlfixer} & 61.5\% & 73.4\% & 69.2\% & 69.1\% & 67.8\% & 10.5\% \\
          & VeriAssist~\cite{huang2024towards} & 65.3\% & 74.8\% & 70.5\% & 71.2\% & 67.1\% & 12.0\% \\
          & AutoVeriFix~\cite{tan2025autoverifix} & 76.4\% & 83.3\% & 79.8\% & 74.3\% & 69.3\% & 14.5\% \\
          \cline{2-8}
          & \cellcolor{softgreen!14}{\textbf{CorrectRTL}} & \cellcolor{softgreen!14}{\textbf{85.2\%} \textcolor{NormalGreen}{\scriptsize(+26.2)}} & \cellcolor{softgreen!14}{\textbf{87.1\%} \textcolor{NormalGreen}{\scriptsize(+17.9)}} & \cellcolor{softgreen!14}{\textbf{85.3\%} \textcolor{NormalGreen}{\scriptsize(+16.7)}} & \cellcolor{softgreen!14}{\textbf{77.4\%} \textcolor{NormalGreen}{\scriptsize(+8.3)}} & \cellcolor{softgreen!14}{\textbf{73.2\%} \textcolor{NormalGreen}{\scriptsize(+6.0)}} & \cellcolor{softgreen!14}{\textbf{22.0\%} \textcolor{NormalGreen}{\scriptsize(+12.5)}} \\
    \midrule\midrule
          \multirow{5}{*}{\shortstack{Claude Sonnet\\4.5~\cite{anthropic2024claude}}} & Direct generation & 80.1\% & 88.8\% & 79.5\% & 68.0\% & 74.0\% & 20.5\% \\
          & RTLFixer~\cite{tsai2024rtlfixer} & 81.4\% & 89.5\% & 80.1\% & 68.0\% & 74.0\% & 22.0\% \\
          & VeriAssist~\cite{huang2024towards} & 83.3\% & 90.2\% & 82.1\% & 70.0\% & 74.0\% & 24.0\% \\
          & AutoVeriFix~\cite{tan2025autoverifix} & 85.3\% & 91.6\% & 85.3\% & 71.0\% & 75.0\% & 27.0\% \\
          \cline{2-8}
          & \cellcolor{softgreen!14}{\textbf{CorrectRTL}} & \cellcolor{softgreen!14}{\textbf{91.3\%} \textcolor{NormalGreen}{\scriptsize(+11.2)}} & \cellcolor{softgreen!14}{\textbf{96.1\%} \textcolor{NormalGreen}{\scriptsize(+7.3)}} & \cellcolor{softgreen!14}{\textbf{90.4\%} \textcolor{NormalGreen}{\scriptsize(+10.9)}} & \cellcolor{softgreen!14}{\textbf{85.0\%} \textcolor{NormalGreen}{\scriptsize(+17.0)}} & \cellcolor{softgreen!14}{\textbf{82.0\%} \textcolor{NormalGreen}{\scriptsize(+8.0)}} & \cellcolor{softgreen!14}{\textbf{49.0\%} \textcolor{NormalGreen}{\scriptsize(+28.5)}} \\
    \midrule\midrule
          \multirow{5}{*}{DeepSeek-V3~\cite{liu2024deepseek}} & Direct generation & 65.5\% & 73.1\% & 63.4\% & 56.0\% & 60.0\% & 13.5\% \\
          & RTLFixer~\cite{tsai2024rtlfixer} & 67.2\% & 75.8\% & 64.1\% & 56.0\% & 60.0\% & 14.5\% \\
          & VeriAssist~\cite{huang2024towards} & 70.1\% & 77.6\% & 67.3\% & 58.0\% & 60.0\% & 16.0\% \\
          & AutoVeriFix~\cite{tan2025autoverifix} & 73.5\% & 80.4\% & 71.2\% & 60.0\% & 62.0\% & 18.5\% \\
          \cline{2-8}
          & \cellcolor{softgreen!14}{\textbf{CorrectRTL}} & \cellcolor{softgreen!14}{\textbf{76.8\%} \textcolor{NormalGreen}{\scriptsize(+11.3)}} & \cellcolor{softgreen!14}{\textbf{82.3\%} \textcolor{NormalGreen}{\scriptsize(+9.2)}} & \cellcolor{softgreen!14}{\textbf{75.1\%} \textcolor{NormalGreen}{\scriptsize(+11.7)}} & \cellcolor{softgreen!14}{\textbf{66.0\%} \textcolor{NormalGreen}{\scriptsize(+10.0)}} & \cellcolor{softgreen!14}{\textbf{68.0\%} \textcolor{NormalGreen}{\scriptsize(+8.0)}} & \cellcolor{softgreen!14}{\textbf{30.0\%} \textcolor{NormalGreen}{\scriptsize(+16.5)}} \\
    \bottomrule
    \end{tabular}
    }
    \footnotesize
    VerilogEval and RTLLM use simulation-based verification; ScaleBench uses JasperGold SEC against golden RTL.
    \\Numbers in parentheses denote improvement over direct generation with the same LLM.
    \caption{Overall functional correctness (Pass@1) across backbones and self-correction methods.}
    \label{tab:overall-rtl-accuracy}
\end{table*}

\textbf{Query formation under spec-level decomposition.}
When Phase~1 uses spec-level decomposition, the per-block RTL instruction $\mathcal{I}_i^\text{R}$ is already a natural-language description of the block's intended behavior. We use $\mathcal{I}_i^\text{R}$ directly as the retrieval query. The retrieved reference code blocks are passed to the RTL agent, which adapts its coding style to the current block. The retrieved SVA templates are passed to the SVA agent, which adapts them to the block's actual signal names and emits one SVA $\mathcal{A}_i$.

\textbf{Query formation under code-level decomposition.}
When Phase~1 uses code-level decomposition, the RTL block $\mathcal{C}_i$ has already been generated, so retrieving more reference code is unnecessary. We still need SVA templates. For each parsed block, an LLM reads $\mathcal{C}_i$ together with the portion of $\mathcal{S}$ that describes it, and emits a short natural-language description of the block's behavior. This description serves as the retrieval query, and only the proven SVA templates attached to the top-$K$ neighbors are returned. The SVA agent then adapts them to $\mathcal{C}_i$'s signal names.

\textbf{RAG for Generations.}
Retrieval boosts both sides of the Phase-2 generation, which leads to a stronger first-pass design before any self-correction is applied.
\emph{(1) RAG for RTL block generation.} The retrieved reference blocks give the RTL agent concrete patterns for the target behavior. The agent adapts a verified implementation instead of composing one from the spec alone, which reduces structural and signal-level errors in the initial code.
\emph{(2) RAG for SVA generation.} Spec descriptions rarely spell out formal assertions, so SVAs generated from scratch often have wrong operators, missing sampling edges, or properties that do not match the intended behavior. The retrieved templates have already been formally discharged on golden RTL. The SVA agent adapts their signal names to the current block, which gives a syntactically well-formed and semantically aligned starting point and removes most of the from-scratch error modes.

\subsection{Functional Correction}\label{sec:functional-correction}

When Phase~3 flags a block $\mathcal{C}_i$, we invoke a rewriter agent with three well-defined inputs: the current RTL block $\mathcal{C}_i$, the violated SVA $\mathcal{A}_i$, and the counterexample log $\xi_i$ emitted by the verifier. The log contains the signal values at the violation cycle and the relevant history needed to explain how the violation was reached. The agent reasons about these three inputs and produces a corrected block $\mathcal{C}_i'$. No retrieval is performed at this step.

\textbf{Analyze-plan-rewrite chain.}
The rewriter is a single LLM call whose structured prompt composes three steps:
\begin{equation}
    \mathcal{C}_i' = \textsc{Rewrite}\!\big(\mathcal{C}_i,\; \textsc{Plan}\!\big(\textsc{Analyze}(\mathcal{C}_i, \mathcal{A}_i, \xi_i),\; \mathcal{A}_i\big)\big).
\end{equation}
The \textsc{Analyze} step produces a root-cause diagnosis $d_i$ in natural language. It traces the counterexample values back through the block body to identify the specific line or expression that makes $\mathcal{A}_i$ fail. The \textsc{Plan} step turns $d_i$ into a concrete modification plan $p_i$ that is expressed in terms of the block's signals and aligned with $\mathcal{A}_i$. The \textsc{Rewrite} step edits $\mathcal{C}_i$ according to $p_i$ and emits the corrected block $\mathcal{C}_i'$. Only the failing block is changed in the full design:
\begin{align}
    \mathcal{C}' = \textsc{Replace}(\mathcal{C},\; \mathcal{C}_i,\; \mathcal{C}_i').
\end{align}
The block-level decomposition from Phase~1 is what makes this targeted replacement safe, since correct blocks remain untouched.

\textbf{Iterative correction.}
After each rewrite, Phases~3 and 4 run again on the updated design $\mathcal{C}'$. Any block that is still flagged is corrected again using the same analyze-plan-rewrite chain on its latest $(\mathcal{C}_i, \mathcal{A}_i, \xi_i)$ triple. The loop stops early once all blocks pass their SVAs, and is capped by a fixed number of rounds to avoid oscillation between rewrites. Syntax correction~\cite{tsai2024rtlfixer} is applied as a preprocessing step and is not discussed further here.

%% file: section/4-experiments.tex
\section{Experiment} \label{sec:expr-setup}

\subsection{Experimental Setup}\label{exp:setup}

\textbf{Evaluation benchmarks.}
We evaluate on four benchmarks spanning different complexity levels: (1) VerilogEval Human v1/v2~\cite{liu2023verilogeval} and VerilogEval Machine v1 (156/156/143 designs, 10--50 LoC), (2) RTLLM v1.1/v2~\cite{lu2024rtllm} (50 designs each, medium complexity), and (3) \textbf{ScaleBench} (200 designs, 30--348 LoC), a more challenging benchmark with balanced lines of code (LoC).

\begin{table*}[t]
\centering\small
\resizebox{0.98\textwidth}{!}{
\begin{tabular}{l|c|c|cc|cc|c}
\toprule
\multirow{2}{*}{\textbf{Decomposition configuration}} &
\textbf{Step 1} &
\textbf{Step 2} &
\multicolumn{2}{c|}{\textbf{Identification (Step 3)}} &
\multicolumn{2}{c|}{\textbf{Correction (Step 3)}} &
\textbf{Final Pass@1} \\
 & Gen Pass@1 & +RAG Pass@1 & Recall & FN & Fix & Regression & Pass@1 \\
\midrule
Vanilla RTL generation        & 20.5\% & ---    & ---  & ---  & 0.0\% & 0.0\% & 20.5\% \\
Code-level decomposition      & 20.5\% & 20.5\% & 42\% & 10\% & 3.5\% & 0.3\% & 23.7\% \\
Vanilla spec decomposition    & 20.5\% & 28.0\% & 50\% & 8\%  & 5.2\% & 0.4\% & 32.8\% \\
\rowcolor{softgreen!14}\textbf{CorrectRTL (Structure-oriented decomposition)} & \textbf{28.0\%} & \textbf{40.0\%} & \textbf{68\%} & \textbf{5\%} & \textbf{9.2\%} & \textbf{0.2\%} & \textbf{49.0\%} \\
\bottomrule
\end{tabular}
}
\caption{Ablation of the four decomposition configurations on ScaleBench (200 designs, Claude Sonnet 4.5).}
\label{tab:paradigm-ablation}
\vspace{-.2in}
\end{table*}

\textbf{Metrics.} Functional \emph{Pass@1} is the primary metric. To isolate the effect of initial code generation and the self-correction step, we compare each design's Pass@1 before and after correction. A design whose Pass@1 goes up contributes to \emph{Fix}. A design whose Pass@1 goes down contributes to \emph{Regression}. Both are reported as averages over the benchmark. The net effect of correction is $\Delta = \text{Fix} - \text{Regression}$.

\textbf{Auxiliary metrics.} Pass@1 is strict at the design level. Many complex designs are hard enough that no method fully passes them, but a corrected design can still be much closer to the golden than its initial code. We therefore report two partial-correctness metrics. \emph{SER} (signal equivalence rate) measures the fraction of output signals that are formally equivalent to the golden. We run JasperGold's sequential equivalence check between the generated RTL and the golden, and parse the log to obtain the per-output proven rate. \emph{SSim} (structural similarity) measures how closely the generated RTL's block structure matches the golden's, as a blend of set overlap and longest-common-subsequence over parsed blocks. A design that misses Pass@1 but lifts SER from, say, $30\%$ to $80\%$ is still a meaningful partial fix. Together, SER and SSim let us quantify progress on designs that remain too hard to pass outright.

We formalize the two metrics as follows. Given a generated RTL $\mathcal{C}$ and its golden reference $\mathcal{C}^\star$:
\begin{align*}
\mathrm{SER}(\mathcal{C}) &= \tfrac{1}{|O|}\,\#\{s \in O : \mathrm{SEC}(s, \mathcal{C}, \mathcal{C}^\star) = \textsc{proven}\}, \\
\mathrm{SSim}(\mathcal{C}) &= \alpha\,\mathrm{Jaccard}\big(B(\mathcal{C}), B(\mathcal{C}^\star)\big) + (1{-}\alpha)\,\mathrm{LCS}\big(B(\mathcal{C}), B(\mathcal{C}^\star)\big),
\end{align*}
where $O$ is the set of output signals, $B(\cdot)$ is the multiset of parsed structural blocks, and $\alpha{=}0.5$.

\textbf{Baselines.} We compare against three recent self-correction methods: RTLFixer~\cite{tsai2024rtlfixer}, VeriAssist~\cite{huang2024towards}, and AutoVeriFix~\cite{tan2025autoverifix}. All baselines run on the same model and temperature as CorrectRTL. Formal verification uses JasperGold 2024.06.

\textbf{Implementation details.} We use the \emph{Microsoft Agentic Framework} to develop our methods and reproduce the baseline methods. We evaluate three backbones: Claude Sonnet 4.5~\cite{anthropic2024claude}, GPT-4o~\cite{achiam2023gpt}, and DeepSeek-V3~\cite{liu2024deepseek}. Generation uses temperature $0.2$ with a single sample.  The RAG retriever uses a pretrained transformer-based sentence encoder (BGE-class, e.g., \texttt{bge-small-en-v1.5}) and a FAISS flat cosine index over the spec-snippet field of $\mathcal{D}$. At inference, we return the top-$K$ neighbors with $K{=}8$. The correction loop caps at $T_{\max}{=}3$ rounds with early stop once all blocks pass.

\subsection{Overall Results}\label{sec:overall-results}

Table~\ref{tab:overall-rtl-accuracy} presents the main results across six benchmarks and three backbone LLMs. CorrectRTL achieves consistent improvements over all baselines on every (model, benchmark) pair.

\textbf{Performance on existing benchmarks in Table~\ref{tab:overall-rtl-accuracy}.} CorrectRTL delivers consistent improvements across the existing benchmarks. With Claude Sonnet 4.5 as the base model, CorrectRTL reaches \textbf{96.1\%} on VerilogEval Machine v1 and \textbf{91.3\%} on Human v1, compared to 91.6\% and 85.3\% for the strongest prior baseline (AutoVeriFix). On RTLLM v1.1, all baseline methods stay consistently below 71\%, while CorrectRTL lifts Claude to \textbf{85.0\%}. This is an improvement of $+17.0$\% over direct generation and $+14.0$\% over AutoVeriFix. These gains show that CorrectRTL remains the strongest method even on benchmarks where prior work is already near saturation.

\textbf{Performance on ScaleBench in Table~\ref{tab:overall-rtl-accuracy}.} The gain from CorrectRTL is larger on the more complex ScaleBench designs. With Claude Sonnet 4.5 as the base model, CorrectRTL improves Pass@1 from 20.5\% to \textbf{49.0\%}, an improvement of $+28.5$\%. The same trend holds for DeepSeek-V3, which CorrectRTL improves from 13.5\% to 30.0\% ($+16.5$\%), and for GPT-4o, which CorrectRTL improves from 9.5\% to 22.0\% ($+12.5$\%). The relative ordering across baselines is preserved on every backbone.

\begin{table}[b]
\vspace{-.2in}
\centering\small
\begin{tabular}{l|cccc}
\toprule
\textbf{Variant} & \textbf{Compile} & \textbf{Correct} & \textbf{Precision} & \textbf{Recall} \\
 & on Step 2 & on Golden & & on fail \\
\midrule
Vanilla RTL Gen    & --- & --- & --- & --- \\
Code-level Decomp  & 55\% & 52\% & 63\% & 42\% \\
Vanilla spec Decomp & 78\% & 70\% & 74\% & 50\% \\
\rowcolor{softgreen!14}\textbf{CorrectRTL} & \textbf{92\%} & \textbf{85\%} & \textbf{88\%} & \textbf{68\%} \\
\bottomrule
\end{tabular}
\caption{Ablation study on SVA quality on ScaleBench.}
\label{tab:sva-quality-by-method}
\end{table}

\subsection{Ablation Study}\label{sec:ablation}

We ablate CorrectRTL along four decomposition configurations that cover the full design space of our method: (1)~\emph{Vanilla RTL generation}, a direct one-shot generation with no decomposition; (2)~\emph{Code-level decomposition}, which generates RTL one-shot and then parses blocks for block-level verification; (3)~\emph{Vanilla spec decomposition}, which splits the specification before generation but without the type-oriented strategies from Sec.~\ref{sec:structure-decomposition}; and (4)~\emph{Structure-oriented decomposition} (ours), which adds the type-oriented strategies on top of spec-level decomposition. Each configuration is evaluated at three pipeline steps: generation (Step~1), RAG-augmented generation (Step~2), and SVA-feedback correction (Step~3).

\begin{figure*}[t]
\centering
\includegraphics[width=0.99\linewidth]{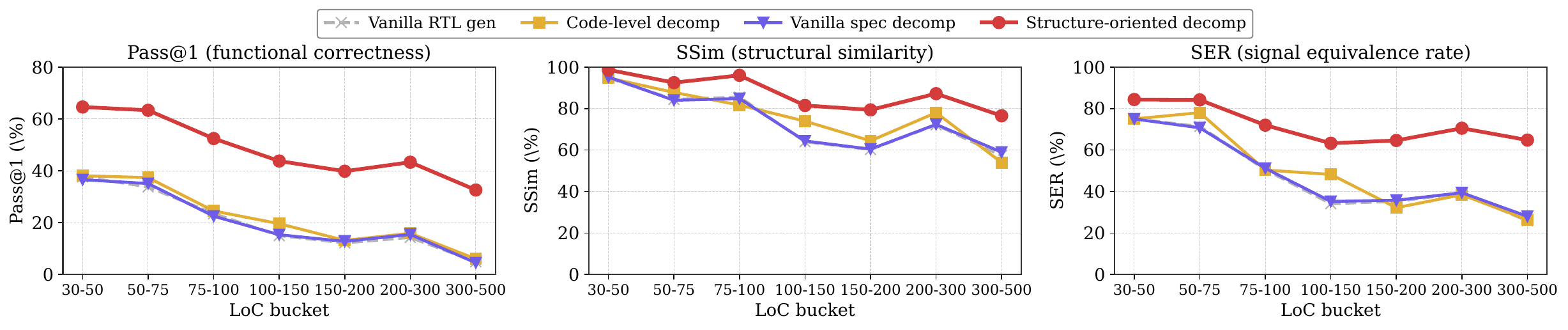}
\vspace{-.2in}
\caption{Pass@1, structural similarity (SSim), and SER by ScaleBench LoC bucket for the four ablation configurations based on the Sonnet 4.5.}
\label{fig:loc-ablation}
\vspace{-.2in}
\end{figure*}

\textbf{Ablation results in Table~\ref{tab:paradigm-ablation}.}
Step~1 Pass@1 stays at 20.5\% for the first three configurations. Code-level decomposition only parses RTL after generation, and vanilla spec decomposition splits the spec without per-type guidance. Neither directly improves the initial RTL. Only the structure-oriented configuration lifts Step~1 to 28.0\%, a $+7.5$\% gain from the type-oriented strategies. Step~2 shows the contribution of RAG. Vanilla RTL generation has no block boundaries, so RAG is not applicable. Code-level decomposition can retrieve SVA templates only, since the RTL is already generated, so Step~2 stays at 20.5\%. Vanilla spec decomposition can retrieve both reference blocks and SVA templates, which lifts Step~2 to 28.0\%. The same RAG step lifts structure-oriented decomposition from 28.0\% to 40.0\%. Vanilla spec decomposition therefore shows a relatively minor effect overall: its Final Pass@1 (32.8\%) is well below structure-oriented decomposition (49.0\%). Step~3 adds SVA-feedback correction. The Final Pass@1 ranges from 20.5\% (vanilla RTL) to 49.0\% (structure-oriented).

\textbf{SVA recall and false-negative rate in Table~\ref{tab:paradigm-ablation}.} SVA recall on Step-2-failing designs is 42\% for code-level, 50\% for vanilla spec, and 68\% for structure-oriented. Vanilla RTL has no block-level SVA, so identification metrics are not applicable. Two factors explain the trend. Code-level SVAs reference internal signal names that do not port across RTL versions, so many fail to compile. Vanilla spec SVAs use spec-mentioned signals but miss domain-specific patterns. Structure-oriented decomposition combines spec-grounded signals with RAG-retrieved proven templates and reaches the highest recall and the lowest false-negative rate. The regression rate stays $\leq$0.4\% across all configurations, so the correction loop rarely overwrites a correct block.

\begin{figure}[b]
\vspace{-.2in}
\centering
\includegraphics[width=0.99\linewidth]{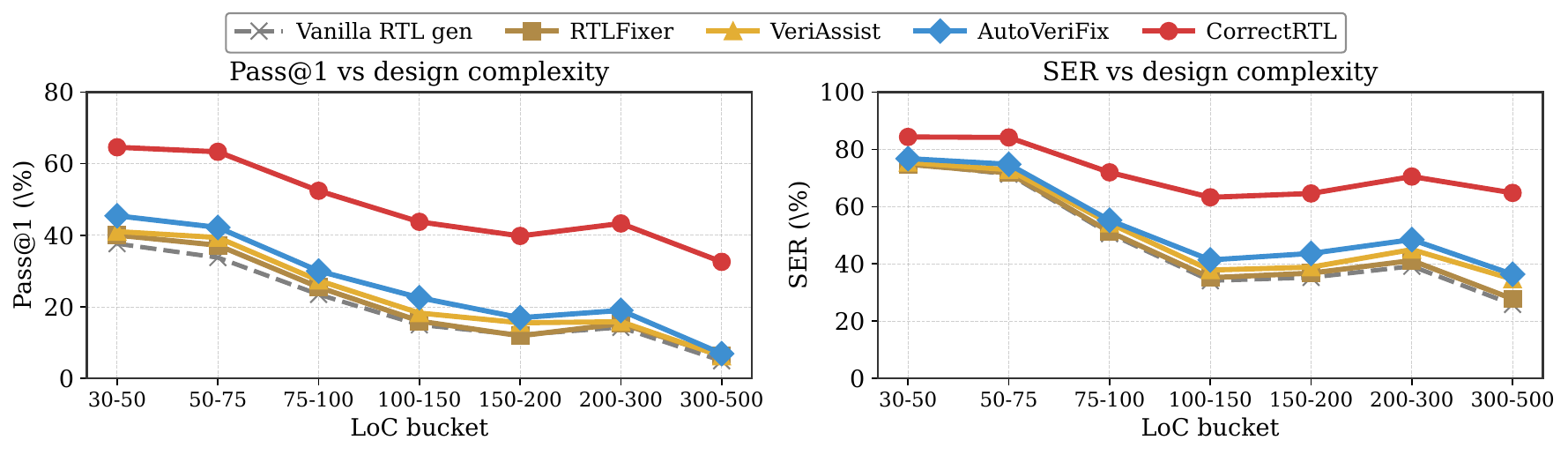}
\caption{Pass@1 (left) and SER (right) versus design complexity (LoC bucket) on ScaleBench for Claude Sonnet 4.5.}
\label{fig:loc-baselines}
\end{figure}

\textbf{SVA quality metrics.} We evaluate each generated SVA set $\mathcal{A}_i$ on four metrics. \emph{Compile on Step~2} is the fraction of SVAs that parse and elaborate against the Step-2 RTL. \emph{Correct on Golden} is the fraction that formally prove on the golden RTL $\mathcal{C}^{\star}_i$, measuring spec-consistency. \emph{Precision} is the fraction of fired SVAs that point to a real bug, computed only over SVAs that fail on a buggy design. \emph{Recall on fail} is the fraction of failing designs in which at least one SVA fires and exposes the bug.

\textbf{SVA quality in Table~\ref{tab:sva-quality-by-method}.}
Code-level SVAs reference code-internal signal names and often fail to compile or prove on the golden. Vanilla spec SVAs use only spec-mentioned signals, so their compile and correctness rates improve, but their recall plateaus without domain-specific patterns. Structure-oriented decomposition paired with SVA-RAG reaches 92\% compile rate, 85\% correctness on the golden, 88\% precision, and 68\% recall on failing designs.

\subsection{Analysis on Design Scale}\label{sec:loc-analysis}

The aggregate Pass@1 numbers above hide how each method behaves as design size grows. ScaleBench spans 30--500 LoC, so we group designs by LoC bucket and report Pass@1, SER, and SSim per bucket. We use this view to compare CorrectRTL with prior baselines and ablation configurations of CorrectRTL.

\textbf{Ablation across design scale in Figure~\ref{fig:loc-ablation}.}
The figure plots Pass@1, SSim, and SER against the ScaleBench LoC bucket for the four ablation configurations. Three observations follow. (i)~SSim stays high and roughly flat (74\%--96\%) across all configurations and all LoC buckets, which suggests that the structural skeleton is recovered well even by weaker configurations. (ii)~Code-level decomposition drops the fastest with LoC. Its Pass@1 falls below 10\% by 150~LoC. Vanilla spec decomposition also drops, though more slowly. (iii)~Structure-oriented decomposition is the only configuration whose Pass@1 stays close to its SSim at large LoC, indicating it preserves both structure and signal-level correctness as designs grow. The SER curves sit between Pass@1 and SSim, consistent with the gap being at the signal level.

\textbf{Comparison with baselines in Figure~\ref{fig:loc-baselines}.}
This figure plots Pass@1 and SER against the ScaleBench LoC bucket for the three baselines and CorrectRTL (Claude Sonnet 4.5). All three baselines drop sharply past 200~LoC. Their Pass@1 falls below 10\% on 200--300~LoC and below 5\% on 300--500~LoC. CorrectRTL drops more slowly with design size: it holds $\geq$30\% Pass@1 through 150--200~LoC and retains 12\% at the largest bucket. The SER curves on the right show the same trend at the signal level. CorrectRTL keeps $\geq$50\% signal-level equivalence on 200--300~LoC designs, where the baselines fall below 35\%. Overall, the gap between CorrectRTL and the baselines widens as designs grow larger.

%% file: section/6-conclusion.tex
\section{Conclusion}\label{sec:concl}

We presented \textbf{CorrectRTL}, a four-phase framework that improves both the generation and the self-correction of RTL designs. We also introduced \textbf{ScaleBench}, a 200-design benchmark of complex RTL with formal equivalence checking against golden references. On ScaleBench, CorrectRTL improves Pass@1 from 20.5\% to 49.0\% with Claude Sonnet 4.5, and consistent gains hold across GPT-4o, DeepSeek-V3, and the small-to-medium VerilogEval and RTLLM benchmarks. We hope this work motivates further research on block-level retrieval and assertion-driven self-correction for large-scale RTL generation.

\section{Acknowledgement}

This work is supported by Hong Kong Research Grants Council (RGC) CRF-YCRG C6003-24Y, 16216825, and T46-415/25-R. It is also supported by RGC JLFS-YSF/E-605/26. It was partially conducted by ACCESS – AI Chip Center for Emerging Smart Systems, supported by the InnoHK initiative of the Innovation and Technology Commission of the Hong Kong Special Administrative Region Government.